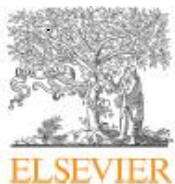

Contents lists available at SciVerse ScienceDirect

## CIRP Annals Manufacturing Technology

Journal homepage: www.elsevier.com/locate/cirp

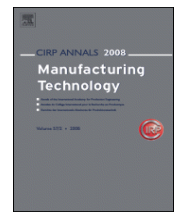

# Knowledge-based multi-level aggregation for decision aid in the machining industry

Mathieu Ritou[1], Farouk Belkadi[2], Zakaria Yahouni[1], Catherine Da Cunha[2], Florent Laroche (2)[2], Benoit Furet[1]

[1] LS2N (Laboratory of Digital Sciences of Nantes, UMR CNRS 6004), University of Nantes, Nantes, France
[2] LS2N (Laboratory of Digital Sciences of Nantes, UMR CNRS 6004), Centrale Nantes, Nantes, France

In the context of Industry 4.0, data management is a key point for decision aid approaches. Large amounts of manufacturing digital data are collected on the shop floor. Their analysis can then require a large amount of computing power. The Big Data issue can be solved by aggregation, generating smart and meaningful data. This paper presents a new knowledge-based multi-level aggregation strategy to support decision making. Manufacturing knowledge is used at each level to design the monitoring criteria or aggregation operators. The proposed approach has been implemented as a demonstrator and successfully applied to a real machining database from the aeronautic industry.

Decision Making; Machining; Knowledge based system

## 1. Introduction

In modern factories, manufacturing activities generate large volumes of digital data, which are partially stored and remain underexploited. The efficient use of manufacturing data is a key leverage point for the management and the improvement of the company performance, notably through a decision-aid system.

Such systems are nowadays based on the analysis of a large volume of heterogeneous data. Such an analysis is impossible manually, so automatic analyses by Data Mining (DM) are required. It consists in using specific algorithms in order to discover interesting hidden information (patterns) in data. The main DM goals are descriptive (e.g. clustering) or predictive (behaviour in the future). More generally, approaches of Knowledge Discovery in Database (KDD) are used, where DM is a particular step in this process. The KDD steps are the data collection, selection, pre-processing, transformation, data mining and, lastly, the interpretation and visualization of the results [1]. This leads to new and useful manufacturing knowledge. KDD approaches are developed in many domains and particularly in manufacturing [2]. As in prognosis techniques, KDD approaches can be classified into three categories: purely data-driven (e.g. statistics or Artificial Intelligence technics), model-based or hybrid [3]. More generally, a model-based approach can be considered as a kind of knowledge-based approach. It can be based on mechanical or empirical models. It can also integrate other types of expert knowledge. To satisfy reliability and interpretability of results, which are the two key points in industry, a combined data-driven and knowledge-based approach is advised. It has been proven that knowledge integration enables a better consideration of expert expectations and of the business context [4]. To do so, existing works addressing the issue of manufacturing knowledge capitalization and reuse can be considered [5,6].

As in every manufacturing domain, there is a growing interest for data mining in the machining industry. Global approaches are especially suitable for flexible production systems. Abundant literature deals with the instrumentation and monitoring of machine-tools [7]. Interesting data can also be collected from the CNC (Computer Numerical Control), through communication protocols such as OPC-UA [8]. Approaches are proposed for the analysis of machining data, based on technics of Machine Learning (ML) and signal processing. The objectives are generally limited to classical monitoring issues at machine level (e.g. estimation of tool wear or surface roughness)[9]. Only a few recent works have expressed interest in global management aspects, such as an application for adaptive scheduling or process improvement [10,11,12]. Furthermore, most of DM or ML works are applied to a database collected during a cutting test campaign in laboratories. Due to the varied situations encountered in industrial flexible production, the complexity of DM is higher, making knowledge-based approach particularly suitable for it.

The analysis of a large volume of manufacturing data leads to Big Data challenges. Indeed, the KDD approaches generally consist in direct and successive computations of the different steps, requiring huge computing power, especially for the DM step [13]. It can rapidly become computationally inefficient to analyse such a large volume of data. One solution is to split the computational problem into smaller ones. Distributed and parallel computing can be envisaged, especially in a context of cloud computing [3]. Another solution is to reduce the volume of data. Technics of dimension reduction (e.g. principal component analysis) are proposed. Data aggregation can be performed at the transformation step of the KDD process. In this way, a smaller volume of simplified and more meaningful data is obtained. Moreover, rather than a unique aggregation, a multi-level approach (i.e. granular computing) is a good solution to address the Big Data problem [14].

The paper presents an original knowledge-based multi-level aggregation approach for data mining in machining. Manufacturing knowledge is integrated at different aggregation steps, from real-time signals to smart data and then KPI. It enables a decrease of data volume while increasing their significance. The approach is tested through a decision-aid demonstrator that is applied to an industrial machining database.

## 2. Decision-Aid Conceptual Framework for machining

By embedding sensors into a machine-tool and connecting them to intelligent engines, new data and knowledge can be generated and interpreted. This knowledge, combined with other know-how and expert rules, is necessary to understand the current status of the machining process and its future changes. Therefore, one significant challenge is how to exploit this digital content for decision making and operational management perspectives. The proposed knowledge-based decision-aid framework provides indicators to support the operational management in the machining industry.

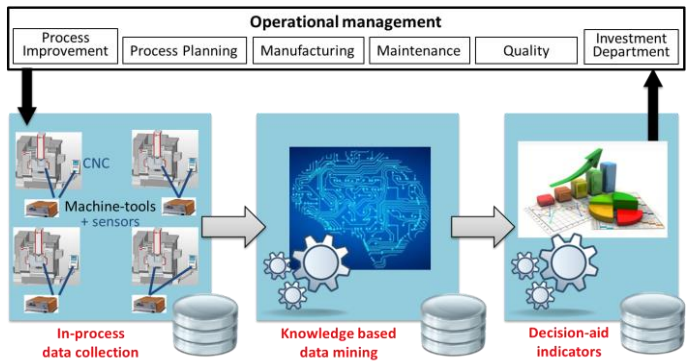

**Figure 1.** Decision–aid framework for machining.

As shown in Figure 1, three software modules are interconnected to support the decision-making process. The first module performs signal processing on sensor measures and collects the in-process heterogeneous data. The next step is a knowledge-based data mining. It uses contextual clustering and unsupervised machine learning algorithms introduced in [12]. Finally, new decision-aid indicators are computed and delivered to various departments according to their needs in terms of process management and improvement. A knowledge-engineering process is used to support the extraction, the formalization and the integration of other manufacturing knowledge.

To be efficient, a decision-aid framework should send the right information at the right moment to decision-makers. To do so, it should be able to manage large quantities of data and ensure flexible circulation of this volume of data throughout the enterprise network. To solve this issue, expert knowledge can be exploited for data aggregation. Only small quantities of meaningful machining-process information are then used for the construction of decision-aid indicators according to the expert intention and the business context. A knowledge-based and multi-level approach is proposed in the next section.

## 3. Knowledge-based multi-level aggregation

Contrary to pure data-driven approaches, the manufacturing knowledge is used in the proposed approach at each level to design the monitoring criteria or aggregation operators (Fig 2). It leads to meaningful information and to more reliable analyses of the production process. Different kinds of knowledge are considered and combined according to the expert point of view, e.g. mechanical model of machining, workpiece properties, business rules, as well as empirical knowledge. In this way, the real-time in-process machining signals lead to smart machining data and then to key performance indicators.

### 3.1. Level 0: real-time signals

A monitoring and data collection system, called EmmaTools, has been developed for machining [15]. Four accelerometers are integrated into the spindle of the machine-tool (in each radial direction, at the front and rear bearings). The signals of in-process machining vibration are measured at a sampling rate of 25 kHz, which is suitable due to the high cutting speeds used in industry. The cutting power is also measured.

### 3.2. Level 1: monitoring and collected data

A record of 4 vibration signals at 25 kHz during months or years of industrial productions would result in an unnecessarily and incommodiously large volume of data. In addition, at each instant, in the frequency domain, only a few vibration components are relevant. Consequently, signal processing in real-time is highly suitable.

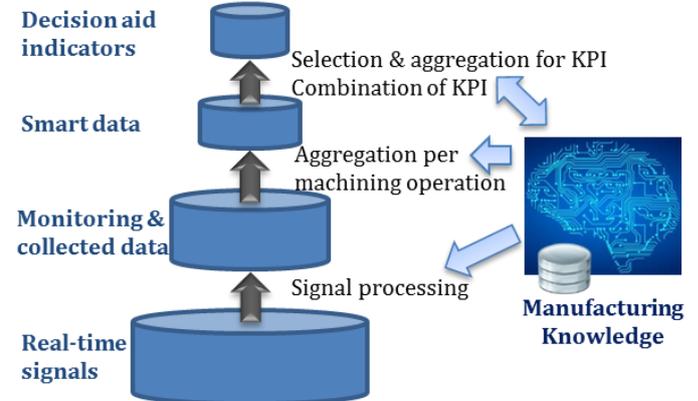

**Figure 2.** Knowledge-based multi-level data aggregation.

The idea is to detect the occurrence of unsuitable specific events (rather than a simple collection of statistical indicators). Dedicated monitoring criteria were designed based on mechanical model of machining, in order to detect tool breakage, spindle failure or chatter (unstable cuts resulting in a too poor surface quality). Some workpiece properties (e.g. material) are also considered as knowledge to tune signal processing parameters (e.g. bandwidth). Order tracking is performed on the vibration signals in the frequency domain. Tool breakage is detected by increased tool unbalance (at spindle frequency), chatter by asynchronous cutting vibration and spindle failure by faulty bearing-induced vibration. Vibration level $V_{RMS}$, a piece of empirical knowledge, is also evaluated. The monitoring criteria are computed every 0.1s. Many machining operational context data (such as machine-tool axes position and speed, or tool and program names) are also collected by the device at a sampling rate of 10 Hz (from the CNC of the machine-tool, by field bus). These operational context data are useful for the following aggregation steps. In this way, collected data (monitoring + operational context data) are synchronous.

### 3.3. Level 2: smart data

On the shopfloor, several workpieces are produced daily on a machine-tool, using several cutting tools for different machining operations. In the aeronautic industry, a machining operation with a given tool can last for several hours, with a large volume of corresponding data. It was chosen to aggregate the monitoring data into few smart data, at the time scale corresponding to one usage period of a given cutting tool. One tool usage period can include several machining operations.

The smart data, related to a given tool usage period, result from an accurate data selection (notably through the operational context data) and from aggregation operators. The latter can be a simple statistic operator e.g. the average power or feedrate during cuts, or the maximum spindle temperature. But these basic operators are not relevant for machining incidents or for the machining vibrations. In fact, machining incidents are very rare (e.g. one severe event for a few seconds per week) and need to be clearly emphasized. It is also known that a moderate vibration level is acceptable, contrary to severe vibration that can damage the tool or the spindle [16]. These are the reasons why a Criticality Operator (*CO*) has been proposed. It quantifies the excessive magnitudes, e.g. of vibrations, over a critical threshold

$T_i$. Let $X_i=\{x_i(k), k=0...n\}$ the time series representing an $i^{th}$ monitoring criterion. The operator is defined as:

$$CO[X_i > T_i] = \sum_{k=ti}^{tf} y(k).dt \quad where \begin{cases} y(k) = x_i(k) - T_i & if\ x_i(k) > T_i \\ y(k) = 0 & if\ x_i(k) < T_i \end{cases}$$

$$T[X_i > T_i] = \sum_{k=ti}^{tf} z(k).dt \quad where \begin{cases} z(k) = 1 & if\ x_i(k) > T_i \\ z(k) = 0 & if\ x_i(k) < T_i \end{cases}$$

with $t_i$ and $t_f$ the initial and final time of the considered period (based on operational context data) and dt = 0.1s the sampling period. A second proposed operator $T$ evaluates the time during which a criterion exceeds a critical threshold. Note that, in the case of vibrations, the monitoring values obtained from each of the four accelerometers are aggregated by a quadratic mean, before the aforementioned aggregation.

The critical thresholds $T_i$ are obtained by unsupervised Machine Learning from the machining database (since it is generally not possible to collect labelled training data in industry, such as the few exact instants where chatter occurred). They become new manufacturing explicit knowledge that can be reused.

The definition of smart data and the way to select and aggregate the data are determined based on business rules and some tacit knowledge of manufacturing experts (issued from their experience). For instance, the smart data for a given tool usage period can be the duration of chatter ($T[Nh>T_{Nh}]$, where $Nh$ is a chatter detection criterion), the critical forced vibration ($CO[V_{RMS}>T_{Vrms}]$), or the average cutting power. This aggregation step is time-consuming, but it can be performed in concurrent operation time. By this means, meaningful and easy-to-use smart data are obtained.

*3.4. Level 3: decision-aid indicators*

The last step is the aggregation of smart data (associated to a given tool-usage period), into Key Performance Indicators (KPI) for decision aid, dedicated to answer the specific managerial needs of different departments of a machining company. A decision-aid indicator is defined as a contextual instantiation of a set of KPI and smart data.

As for smart data, the definition of KPI and the way to select and aggregate smart data are determined based on manufacturing knowledge such as business rules and best practices. For example, the definition can be the sum or mean (weighted or not) of contextually selected smart data. It can also take the form of a comparison with the usual value of a given performance indicator that has been previously assessed by machine learning. Examples of KPI are provided in the use case (next section). Thereby, models defining KPI and smart data for machining constitute new formal knowledge.

The instantiation of KPI models rely on accurate data selection, which is enabled by operational context data. The definition of a decision-aid indicator is obtained according to four instantiation parameters, which describe the context of the decision:
- the objective of the decision for which decision-aid indicators are defined;
- the decider who exploits the resulting decision-aid indicator according to a given business point of view;
- the scope of analysis in terms of machine, workpiece, program, tool, or other entities;
- the instantiation mode that indicates when a given decision-aid indicator is computed. Three modes are considered: periodic, on demand, and on event.

**4. Industrial use case**

For validation purpose, the decision-aid system has been firstly tested with 2 well-known industrial use cases. The present section details the data exploitation scenarios. The industrial use cases concern the high speed machining of structural aeronautic parts in aluminum alloy. These high added-value parts present thin walls and floors, subject to vibration issues (the main cause of non-quality, requiring a costly manual polishing on the part). The production system is also very flexible: a machine-tool almost never produces the same part twice consecutively. Consequently, a global approach is suitable for decision aid.

*4.1. Application scenario: reporting for Manufacturing Dpt*

A reporting scenario was conducted as an example of a decision-aid strategy. The reporting strategy is based on the instantiation parameters described in Section 3.4. The role of the reports is to provide useful knowledge to the manufacturing department, which enables a better understanding of the production process. Two examples were chosen for the demonstration in the aim to solve the chatter problems (unstable cuts that result in unacceptable surface roughness on the workpiece). The objective of the first one is to identify the worst cutting tools (for which the cutting conditions should be optimized). The second one highlights the main faulty workpiece programs (requiring updates). The reporting strategy follows a periodic mode, where decision-aid indicators are evaluated every 3 months.

To support the demonstration, an EmmaTools monitoring device was installed on a machine-tool of an aeronautic company. 39 Gb database of monitoring and operational context data were collected in-process, over 426 days of industrial production, where 80 cutting tools and 346 programs machined 534 workpieces. The volume of the database is reasonable (approximately 100 Mb of machining data per machine-tool and per day), thanks to the real-time processing of accelerometer signals that computes monitoring criteria, such as $Nh$ for chatter detection. The second step aggregates the monitoring data into smart data, based on the chatter duration rule: $T[Nh>20m/s^2]$ for each tool usage period. It leads to a total volume of smart data of approximately 2.4 Mb per year and per machine-tool. The third aggregation step results in a decision-aid indicator as a combination of the KPIs for tool and program. The KPIs are evaluated through the total duration of chatter per tool and per program respectively, over 3 months of production.

*4.2. Demonstrator results*

To implement this scenario, a software demonstrator was developed for decision aid, based on the in-process machining data, and applied to the database collected in industry. Multi-Agent technology is used for information management between software modules composing the proposed decision-aid framework [17]. An agent is usually defined as a virtual autonomous entity that perceives the environment and acts upon an environment. The originality of the proposed approach is to adapt the concept of multi-agent to support interoperability and software interactions management. In fact, the communications are only between agents or between an agent and a software module, while the different software modules have no connection between them.

In the reporting scenario, four agents are used to monitor the execution of software components. The first agent is a Human-Machine Interface (HMI) that enables interaction with users. The Traceability agent aims to manage the connection to various data and knowledge bases. The third category is called a Computing agent and supports the remote execution of various Matlab programs in charge of data aggregation and performance indicator evaluation. The last category is the Reporting agent which is in charge of the definition and sharing of reports based on the four instantiation parameters of the reporting strategy.

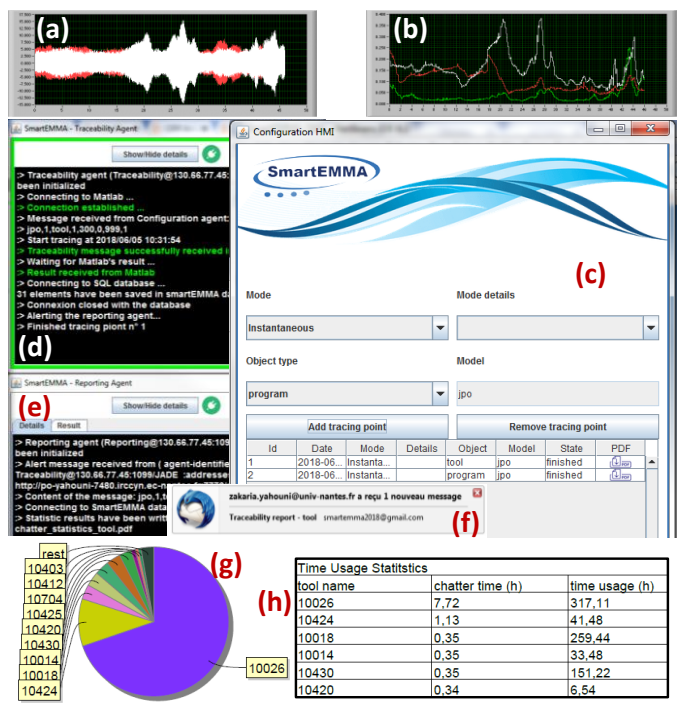

**Figure 3.** Demonstrator results for the reporting scenario.

The heterogeneous agents interact in order to carry out the reporting process. Firstly, the decider is connected to the HMI agent for the configuration of the traceability and reporting parameters. Based on the request, the Traceability agent identifies in the repositories all data and knowledge required for the construction of decision-aid indicators. When the monitoring data are extracted from the database, the Computing agents manage the smart data aggregation and the KPI assessment. The Traceability agent stores the results. Finally, the Reporting agent extracts the computed indicators and generates a PDF report that presents the results in the appropriate format depending on the profile of the decider.

Figure 3 shows the demonstrator interfaces of the decision-aid process for the example of cutting tool scenario (defined in Section 4.1). The accelerometer signals (Fig. 3a) are processed in real time, obtaining monitoring criteria that are collected (e.g. $V_{RMS}$, Fig. 3b). The smart data are computed in concurrent operation time. The HMI (Fig. 3c) enables the configuration of the reporting parameters, notably by selecting the traceability objective, the scope (program, tool, etc.) and the instantiation mode. The computations of KPIs are then carried out. The real time execution of the different agents is shown in Fig. 3d and 3e. It results in the automatic sending of an e-mail to the decider with the appropriate PDF report (Fig. 3f). The report contains a visualization of the decision-aid indicator, here a pie chart of the chatter statistics per tool (Fig. 3g). It indicates that the tool reference 10 026 causes most of the chatter occurrences and needs an optimization of its cutting conditions. Fig. 3h provides more details about the summed chatter durations per tool that are visualized in the pie chart.

Since two well-known application scenarios had been chosen, the results were compared with the experts' expectations and have shown consistency. In this way, the quality of the system was validated.

## 5. Conclusion

This paper introduces an original knowledge-based and multi-level approach for data management and mining in machining. The different data aggregation steps are conducted with the integration of manufacturing knowledge, in order to pass from real-time signals to smart data, KPI and lastly to contextual decision-aid indicators. This strategy progressively reduces the volume of data and increases their meaningfulness.

The approach was implemented in a demonstrator of the decision-aid system. The architecture is based on a multi-agent system. An example of reporting scenario for the manufacturing department was successfully implemented. The related KPI were proposed and evaluated on a real machining database from the aeronautic industry. It led to the automatic sending of an e-mail, with the resulting decision-aid indicator, to the decider. The different levels of knowledge-based aggregated data are illustrated in the example. The efficiency of the proposed approach to tackle the big data issue was shown. Furthermore, the positive feedback from the industrial partners of the project provided a first validation.

More complex decision-aid scenarios are under development, concerning the analysis and the diagnosis of critical events and the prediction of failure.

## 6. Acknowledgement

The authors acknowledge the financial support of the French government of the ANR SmartEmma project (ANR-16-CE10-0005) and they would also like to thank the contributions made by the industrial partners.